# Parameter adjustment in Bayes networks.
# The generalized noisy OR–gate

**F. J. Díez**

Dpto. Informática y Automática. UNED

Senda del Rey. 28040 Madrid. Spain

<FJavier.Diez@human.uned.es>

## Abstract

Spiegelhalter and Lauritzen [15] studied sequential learning in Bayesian networks and proposed three models for the representation of conditional probabilities. A forth model, shown here, assumes that the parameter distribution is given by a product of Gaussian functions and updates them from the $\lambda$ and $\pi$ messages of evidence propagation. We also generalize the noisy OR–gate for multivalued variables, develop the algorithm to compute probability in time proportional to the number of parents (even in networks with loops) and apply the learning model to this gate.

## 1   INTRODUCTION

Knowledge acquisition is one of the bottlenecks in expert systems building. Bayesian networks (BNs) [10, 11, 8], besides having theoretical grounds supporting them as the soundest framework for uncertain reasoning, offer an important advantage for model construction. In this field, the task consists of building a network which, based on the observed cases, gives the optimal prediction for future ones. Fortunately, in this field there is a normative theory for learning, provided again by Bayesian analysis: we look for the most probable BN given the observed cases.

Once we have chosen the variables, the first step is to determine the qualitative relations among them by drawing causal links. Then, quantitative information must be obtained in the form of conditional probabilities.

The optimal situation happens when there is an available database containing a large number of cases in which the values of all variables are specified. Efficient algorithms for eliciting both the structure and the parameters of the network have been developed. This process performed on a database is called *batch learning*. It allows the automated discovery of dependency relationships. Recent work on batch learning can be found in [4].

Unfortunately, if the cases contained in the database are incomplete, the mathematical framework becomes complex and no definitive theory of learning is established yet. Very often, there is not even an incomplete database, and the knowledge engineer must resort to subjective assessments of probability obtained from human experts, who make use of their memory and of the literature published in their field. In this second situation, the model needs refinement, and it is desirable to endow the system with some capability of adaptation as it executes. This process is called *sequential learning*.

The most important work on sequential learning in BNs was performed by Spiegelhalter and Lauritzen (S–L) [15]. They introduce some assumptions, mainly global and local independence of parameters (see next section), in order to make the problem tractable. Nevertheless, the problem is still difficult when data are incomplete, i.e. when not all the variables are determined. A final problem is the representation of conditional probability tables $P(x|pa(x), \theta_x)$; S–L propose three different models: discretization of parameters, Dirichlet distributions, and Gaussian distributions for the log-odds relative to the probability of a reference state. The second approach is applied in [9] and [14].

The problem addressed in this paper is, as in S–L, to update sequentially the parameters of a probability distribution. The main difference from their work is that we assume a normal distribution for the parameters (not for the log-odds). We will first study the general case and then the OR–gate.

The present work can be important for three reasons:

— Usually, only some nodes in a BN are observable. This means that we do not have databases where every variable is instantiated, and even if we had, we should take them with criticism, remembering that the value stored for an unobservable or unobserved variable was not obtained directly but inferred from the values of other variables. Unfortunately, the construction of general BNs from incomplete databases is very com-plex [1] and no normative algorithm exists yet. So we assume that an initial causal network has been elicited from human experts. The use of Gaussian distributions allows us to integrate easily subjective assessments ("the probability is about 60 or 70%") and experimental results ("$U$ produces $X$ in 67±4% of the patients"). Parameter adjustment takes place when the network performs diagnosis ·on new cases. Our model of learning can naturally deal with incomplete and uncertain data.

— In general, the number of parameters required for every family is exponential in the number of parents, and so is the time for evidence propagation. In the OR–gate, on the other hand, the number of parameters is proportional to the number of causes. This difference can be considerable in real-world applications, such as medicine, where there are often more than a dozen known causes for a disease. When building a BN from a database, the resulting model for the OR–gate will be more accurate if there are only a few cases for every instantiation of the parent nodes. Also for a human expert, it is much easier to answer a question like "What is the probability of $X$ when only cause $U$ is present?" than lots of questions entailing a complex casuistry. It is useful, then, to have a model of sequential learning for that gate.

— The OR–gate is not only valuable for knowledge acquisition, but also for evidence propagation. When applied instead of probability tables, it can save an important amount of storage space and processing time. However, algorithms usually employed for probabilistic inference do not take advantage of this possibility. Section 3.1 generalizes the noisy OR to multivalued variables and develops efficient formulas for propagating evidence. They allow the local conditioning algorithm [2] to exploit the OR–gate even in multiply-connected networks.

# 2  PARAMETER ADJUSTMENT

## 2.1  ASSUMPTIONS

We introduce in this section the hypotheses which constitute the basis of our model. Every case $i$ is given by the instantiation of some variables corresponding to the observed evidence, $e_i$.

**Assumption 1 (Cases independence)** *Cases* $e_i$ *are independent given the parameters:*

$$P(e_1, \ldots, e_N | \Theta) = \prod_{i=1}^{N} P(e_i | \Theta). \qquad (1)$$

This assumption seems reasonable: the probability of a new case depends only on the parameters of the model (the conditional and *a priori* probabilities), not on the cases we have found so far or are going to find in the future. This assumption is the key for the sequential

updating of probabilities, namely

$$P(\Theta | e_1, \ldots, e_N) = \alpha \, P(e_N | \Theta) \cdot P(\Theta | e_1, \ldots, e_{N-1}) \qquad (2)$$

where $\alpha = P(e_1, \ldots, e_{N-1}) / P(e_1, \ldots, e_N)$ is a normalization factor independent of $\Theta$.

**Assumption 2 (Parameter independence)** *All parameters are a priori pairwise independent.*

This includes *local independence* (within every family formed by a node and its parents) as well as *global independence* (among different families). It is not clear that it holds for an arbitrary election of parameters. S–L show the case of pedigree analysis as an example in which global independence is clearly violated.

Both these assumptions where introduced in S–L. The specific feature of our model is as follows:

**Assumption 3 (Gaussian distributions)** *The initial distribution for every parameter $\theta_i$ is given by a Gaussian distribution*

$$P(\theta_i | e_1, \ldots, e_{N-1}) = \mathcal{N}(\mu_i, \sigma_i^2) \qquad (3)$$

*with*

$$0 < \mu_i < 1 \qquad (4)$$

*and*

$$\sigma_i \ll \min(\mu_i, 1 - \mu_i). \qquad (5)$$

Eq. (5) implies that $f(\Theta) \approx 0$ outside the interval $[0, 1]^{N_\Theta}$. This assumptions will allow us to apply the simplifications derived below.

## 2.2  STATISTICAL PROPERTIES

We start now from a multivariate normal distribution of uncorrelated variables. It can be represented as the product of Gaussian distributions:

$$f(\Theta) = \prod_i \mathcal{N}(\mu_i, \sigma_i) \qquad (6)$$

$$= \prod_i \frac{1}{\sqrt{2\pi}\sigma_i} \exp\left[-\left(\frac{\theta_i - \mu_i}{\sigma_i}\right)^2\right] \qquad (7)$$

and study a new distribution given by

$$f'(\Theta) = \begin{cases} c\left(a + \sum_i b_i\theta_i\right) f(\Theta) & \text{for } 0 \leq \theta_i \leq 1, \forall i \\ 0 & \text{otherwise} \end{cases} \qquad (8)$$

where $c$ is a normalization constant. We shall assume, without loss of generality, that it is positive. Then,

$$a + \sum_i b_i\theta_i \geq 0 \quad \text{if } 0 \leq \theta_i \leq 1, \forall i \qquad (9)$$

so that the distribution is always non-negative.

Since conditions (4) and (5) guarantee that $f(\Theta) \approx 0$ outside the interval $[0, 1]^{N_\Theta}$, we have very approximately

$$c = a + \sum_i b_i\mu_i. \qquad (10)$$

With the same approximation, the moments for the new distribution are

$$\mu_i' = E'(\theta_i) = \mu_i + \Delta_i \tag{11}$$

$$\sigma_i'^2 = E'(\theta_i^2) - \mu_i'^2 = \sigma_i^2 - \Delta_i^2 \tag{12}$$

where

$$\Delta_i \equiv \frac{\sigma_i^2 \, b_i}{a + \sum_j b_j \mu_j} \ . \tag{13}$$

(Observe that the normalization constant $c$ is irrelevant. This is an advantage for the computations in parameter adjustment, because it makes unnecessary to normalize messages.)

The covariance will be given by

$$\mathrm{cov}(\theta_i, \theta_j) = E'(\theta_i \theta_j) - \mu_i' \mu_j' = \Delta_i \Delta_j \ . \tag{14}$$

Properties (4) and (5), together with the condition of non-negativity (9), allow us to conclude that $\Delta_i < \sigma_i$, so that, as expected, $\sigma_i'^2 > 0$. We also observe that the standard deviation will be reduced when $b_i \neq 0$. This property will ensure the convergence when applying this study to parameter adjustment.

## 2.3 ALGORITHMS

The purpose of this section is to apply the statistical model in order to update the parameters according to eq. (2). The set of parameters can be partitioned into three subsets relative to an arbitrary variable $X$ (see fig. 1): $\Theta_X$ includes the parameters relative to family $X$, i.e. the parameters that determine $P(x|u)$, where $u$ represents any of the possible configurations for the states of the parents of $X$; $\Theta_X^+$ and $\Theta_X^-$ represent the parameters determining the probabilities in the *families* above the parents of $X$ or below $X$, respectively. In the same way, $\Theta_{UX}^+$ represents the parameters above link $UX$. The new case we are considering consists of the evidence $\mathbf{e}_N$ observed. What we have is $\pi(x)$ and $\lambda(x)$ for every node $X$ and we are going to update the parameters according to this information. The conditional probability is given by

$$P(\mathbf{e}_N|\Theta) \propto \sum_x \lambda(x|\Theta_X^-)\, \pi(x|\Theta_X, \Theta_X^+) \tag{15}$$

$$= \sum_x \sum_u \lambda(x|\Theta_X^-)\, P(x|u, \Theta_X) \prod_i \pi_X(u_i|\Theta_{U_i X}^+)\ . \tag{16}$$

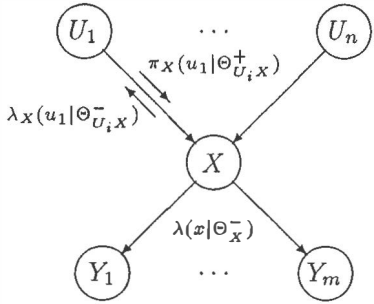

Figure 1: Messages for family $X$.

The definitions of $\pi$ and $\lambda$ are taken from [11].

These formulas are general and can be applied to the case of complete conditional probability tables as well as to the AND/OR–gates.

We first study the general case, represented by giving a parameter for every instantiation of family $X$: $P(x|u, \Theta_x)$. But the probabilities for all values of $X$ must sum up to 1. So they can be represented by the following independent parameters $\theta_x^{\mathbf{u}}$:

$$P(x|u, \Theta_x) = \begin{cases} \theta_x^{\mathbf{u}} & \text{for } x \neq x_0 \\ 1 - \displaystyle\sum_{x' \neq x_0} \theta_{x'}^{\mathbf{u}} & \text{for } x = x_0 \end{cases} \tag{17}$$

With these parameters, eq. (16) turns into

$$P(\mathbf{e}_N|\Theta) \propto \lambda(x_0|\Theta_X^-) +$$

$$\sum_{x \neq x_0} \sum_u [\lambda(x|\Theta_X^-) - \lambda(x_0|\Theta_X^-)]\, \theta_x^{\mathbf{u}} \prod_i \pi_X(u_i|\Theta_{U,X}^+). \tag{18}$$

From the axioms in section 2.1, we had

$$P(\Theta|\mathbf{e}_1, \ldots, \mathbf{e}_N) = P(\mathbf{e}_N|\Theta) f(\Theta_X^-) f(\Theta_X) \prod_i f(\Theta_{U_i X}^+) \tag{19}$$

where every $f$ represents a product of univariate Gaussian distributions.

The last two expressions can be combined and, after integration, we get

$$P(\Theta_X|\mathbf{e}_1, \ldots, \mathbf{e}_N) \propto$$

$$\left[ \hat\lambda(x_0) + \sum_{x \neq x_0} \sum_u [\hat\lambda(x) - \hat\lambda(x_0)]\, \theta_x^{\mathbf{u}} \prod_i \hat\pi_X(u_i) \right] f(\Theta_X) \tag{20}$$

where

$$\hat\lambda(x) \equiv \int \lambda(x|\Theta_X^-)\, f(\Theta_X^-)\, d\Theta_X^- \tag{21}$$

$$\hat\pi_X(u_i) \equiv \int \pi_X(u_i|\Theta_{U_i X}^+)\, f(\Theta_{U_i X}^+)\, d\Theta_{U_i X}^+. \tag{22}$$

Let us also define

$$\hat P(x|u) \equiv \begin{cases} \mu_x^{\mathbf{u}} & \text{for } x \neq x_0 \\ 1 - \displaystyle\sum_{x' \neq x_0} \mu_{x'}^{\mathbf{u}} & \text{for } x = x_0 \end{cases} \tag{23}$$

and

$$\hat\pi(x) \equiv \sum_u \hat P(x|u) \prod_i \hat\pi_X(u_i)\ . \tag{24}$$

By comparison of eq. (20) to the study in section 2.2 and applying the equivalence

$$\hat{\lambda}(x_0) + \sum_{x \neq x_0} \sum_u [\hat{\lambda}(x) - \hat{\lambda}(x_0)] \mu_x^u \prod_i \hat{\pi}_X(u_i)$$
$$= \sum_x \hat{\lambda}(x) \, \hat{\pi}(x) \, ,$$

we eventually find

$$\Delta_x^u = \frac{(\sigma_x^u)^2 \, [\hat{\lambda}(x) - \hat{\lambda}(x_0)] \prod_i \hat{\pi}_X(u_i)}{\sum_{x'} \hat{\lambda}(x') \, \hat{\pi}(x')} \, . \qquad (25)$$

In conclusion, the new distribution for parameter $\theta_x^u$ has changed its mean value from $\mu_x^u$ to $\mu^x + \Delta_x^u$ and its variance from $(\sigma_x^u)^2$ to $(\sigma_x^u)^2 - (\Delta_x^u)^2$. Nevertheless, we do not have Gaussian distributions any more and, in general correlations arise when the case is not complete, i.e. when the observed evidence is not sufficient to determine exactly the values of all variables.

But in order to have a tractable model, we shall *assume* that the new distribution can be approximated by a product of Gaussian univariate distributions,

$$P'(x|u, \Theta_X) = \begin{cases} \mathcal{N}(\mu_x^u + \Delta_x^u, \sigma_x^{u2} - \Delta_x^{u2}) & x \neq x_0 \\ 1 - \sum_{x^* \neq x_0} P'(x^*|u, \Theta_x) & x = x_0 \end{cases} \qquad (26)$$

so that case $\mathbf{e}_{N+1}$ can be treated in the same way, thus having a sequential framework for learning.

## 2.4  COMMENTS

• The approximation in eq. (26) is valid when $\Delta_x^u$ is small compared to $\min(\mu_x^u, 1 - \mu_x^u)$. Otherwise the resulting distribution will differ from a Gaussian function and, besides, correlations given by eq. (14) will not be negligible if standard deviations are wide and observed values were *a priori* improbable. Therefore, those approximations are justified when $\sigma_x^u$ is small, that is to say, when the original model is relatively accurate.

• Messages $\hat{\lambda}(x)$, $\hat{\pi}_X(u_i)$ and $\hat{\pi}(x)$ can be obtained locally, i.e. considering the messages received at node $X$ and the parameters of its family. This is a consequence of the *global independence* assumption. It allows a distributed learning capability (see [3] and fig. 2).

• Eq. (23), for $\hat{P}(x|u)$, is equivalent to eq. (17) for $P(x|u, \Theta_x)$. The only difference is that average values must be taken instead of the original distribution. The same is true for $\hat{\pi}(x)$ in eq. (24). Therefore, evidence propagation in this model is formally equivalent to the "traditional" case, by using mean values instead of exactly determined probabilities. In other words, we need not worry about distributions: we take the average value of each parameter and can neglect, for the moment, the standard deviation.

• According to eq. (25), $\Delta_x^u = 0$ when $\sigma_x^u = 0$. Naturally, a parameter will not be updated if it is exactly determined.

• We observe that $\Delta_x^u = 0$ when $\hat{\lambda}(x) = \hat{\lambda}(x_0)$. As expected, parameters of a family are not updated unless some evidence arrives from its effects. In case $\hat{\lambda}(x) \neq \hat{\lambda}(x_0)$ and $\sigma_x^u \neq 0$, then $\Delta_x^u \neq 0$, at least for some values of $x$ and $u$. According to eq. (12), the standard deviation of a parameter is reduced each time evidence is observed for its corresponding configuration state.

• Every node without parents has an *a priori* probability, which can be dealt with as an ordinary conditional probability by adding a dummy node representing a fictitious binary variable whose value is always TRUE.

• The equations derived in this section, including eq. (25), do not change even if some $\hat{\lambda}$ or $\hat{\pi}$ is multiplied by a constant. It is not necessary to have normalized $\pi$'s, and instead of defining

$$\pi(x|\Theta_X, \Theta_X^+) \equiv P(x|\mathbf{e}_X^+, \Theta_X, \Theta_X^+), \qquad (27)$$

after [10], we could have defined it after [13, 2]:[1]

$$\pi(x|\Theta_X, \Theta_X^+) \equiv P(x, \mathbf{e}_X^+|\Theta_X, \Theta_X^+). \qquad (28)$$

Therefore, this formalism can also be applied when evidence is propagated using the local conditioning algorithm [2] and so the learning method can be applied to general networks as well as to singly-connected ones.

## 3  THE GENERALIZED NOISY OR–GATE

### 3.1  DEFINITION AND ALGORITHMS

The noisy OR–gate was introduced in [10]. In this model, a parent node of $X$ is not conceived as a mere factor (age of the patient, for instance) modulating the probability of $X$ given a certain configuration of the other parents (sex, weight, smoking, etc.). Instead, node $X$ represents a physical-world entity (for example, a disease) that may be present or absent, and its parents represent phenomena —in general anomalies— whose presence can produce $X$. In other words, a link in the OR–gate represents *the intuitive notion of causation* ("$U$ produces $X$"), not only the statistical definition given in [12].

The main advantage of the OR–gate is that the number of parameters is proportional to the number of causes, while it was exponential in the general case. As a consequence, the OR–gate simplifies knowledge acquisition, saves storage space and allows evidence propagation in time proportional to the number of parents.

A generalization for multivalued variables was introduced by Henrion [5] in order to simplify knowledge acquisition. This model can also save storage space, but if a clustering method is chosen for evidence propagation, the conditional probability table of every family must be worked out in advance [7, 8], thus wasting the computational advantage of the OR–gate. For this reason, after formalizing the model, we will now develop an algorithm for computing probability in time proportional to the number of causes, which can also deal with multiply-connected networks.

---

[1] Only eqs. (18) and (20) would be slightly modified. We have here chosen the original definition just for simplicity.

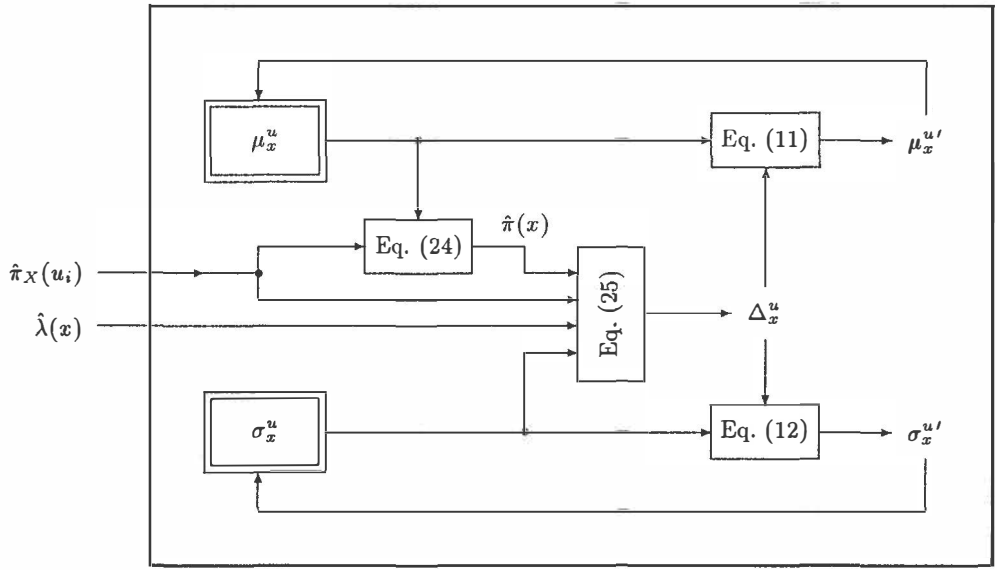

Figure 2: Learning at node $X$.

**Definition 1 (Graded variable)** *A variable $X$ that can be either absent, or present with $g_X$ degrees of intensity, is said to be a graded variable. It has $g_X + 1$ possible values, which can be assigned non-negative integers such that $X = 0$ means "absence of $X$" and succeeding numbers indicate higher intensity.*

Observe that the concept of "graded" is independent of the number of outcomes, since not all multivalued variables represent different degrees of presence and, conversely, the definition includes also the binary variables intervening in the noisy OR, which are of type absent/present $(g = 1)$ and differ from non-graded binary variables, such as sex. The concept is relevant because *the OR-gate makes sense only for graded variables.*

The parameters for a certain family in the OR-gate will be the conditional probabilities of $X$ given that all causes but one are absent; in case $U$ is a cause of $X$, and $V$ agglutinates all the other causes, we have:

$$\theta_{X=x}^{U=u} \equiv P(X = x | U = u, V = 0) \qquad (29)$$

which could be abbreviated as

$$\theta_x^u \equiv P(x | u, v_0) \qquad (30)$$

Obviously

$$\theta_{x_0}^u = 1 - \sum_{x=1}^{g_X} \theta_x^u. \qquad (31)$$

When $U$, as well as all other causes, is absent, $X$ must be absent too:[2]

$$\theta_x^{u_0} = \begin{cases} 1 & \text{for } x = 0 \\ 0 & \text{for } x \neq 0. \end{cases} \qquad (32)$$

In consequence, only $g_U \cdot g_X$ parameters are required for this link.

We now introduce the following definition:

$$Q_U(x) \equiv P(X \leq x | e_{UX}^+, V = 0), \qquad (33)$$

which is the probability of $X$ given all the evidence above link $U \rightarrow X$ in the case that all other causes of $X$ were absent. From

$$P(x | e_{UX}^+, v_0) = \sum_u P(x | u, v_0) P(u | e_{UX}^+)$$

$$= \sum_u \theta_x^u \pi_X(u), \qquad (34)$$

it can be deduced that

$$Q_U(x) = 1 - \sum_{u=1}^{g_U} \pi_X(u) \sum_{x'=x+1}^{g_X} \theta_{x'}^u. \qquad (35)$$

So far in this section we have introduced only some definitions and no assumption. Now we are going to present the key axiom of the OR–gate, which will allow us to calculate the probability of $X$ when more than one of its causes are present.

---

[2]The leaky probability [5] can be assigned to a certain anonymous cause. In this way, the treatment is trivially extensible to the leaky OR–gate.

**Definition 2 (Generalized noisy OR–gate)** *In a Bayes network, given a graded variable $X$ with parents $U_1, \ldots, U_n$ (also graded variables), we say that they interact through a generalized noisy OR–gate when*

$$P(X \leq x | u_1, \ldots, u_n)$$
$$= \prod_i P(X \leq x | U_i = u_i, U_j = 0, j \neq i). \quad (36)$$

The interpretation is as follows: the degree reached by $X$ is the maximum of the degrees produced by the causes acting independently, without synergy [5]. So eq. (36) reflects the fact that $X \leq x$ only when every cause has raised $X$ to a degree not higher than $x$. This model of interaction could also be termed MAX–gate. In the same way, the graded AND–gate could be called MIN–gate.

With definition

$$Q(x) \equiv P(X \leq x | \mathbf{e}_X^+), \quad (37)$$

it is straightforward to get $\pi(x)$ from $Q(x)$:

$$\pi(x) = \begin{cases} Q(x) - Q(x-1) & \text{for } x \neq 0 \\ Q(0) & \text{for } x = 0. \end{cases} \quad (38)$$

According to eq. (33), we have

$$Q(x) = \prod_i Q_{U_i}(x), \quad (39)$$

which allows us to compute $\pi(x)$. To summarize, from $\pi_X(u)$ we get $Q_{U_i}(x)$, and combining all these messages, we arrive through $Q(x)$ at $\pi(x)$ in time proportional to the number of causes, as claimed before.

In case family $X$ formed part of a loop, local conditioning [2] should be applied; then, $\pi$-messages are not normalized, but these formulas remain valid with minor modifications. Moreover, if only link $U_i X$ lies in the loop path, conditioning does not apply to other $Q_{U_j}(x)$ messages, and this allows an important additional save in computation for the OR–gate.

A similar treatment could be made for the AND–gate; we have studied the OR–gate because it appears much more often. An additional advantage of these gates is that they enable us to generate explanations of why the evidence at hand has increased or reduced the probability of $X$ [6].

## 3.2    PARAMETER ADJUSTMENT FOR THE OR–GATE

We are now going to develop a formalism for parameter adjustment in the OR–gate, similar to that of section 2.3 for the general case. The starting point is eq. (15). The expression for $\pi(x | \Theta_X, \Theta_X^+)$ is similar to eq. (38), just including the conditioning on the parameters $\Theta$. In the same way, the expression for $Q(x | \Theta_X, \Theta_X^+)$ is similar to eq. (39); now, global independence of parameters leads to

$$Q(x | \Theta_X, \Theta_X^+) = \prod_U Q_U(x | \Theta_X^U, \Theta_{UX}^+), \quad (40)$$

with $\Theta_X^U$ being the parameters associated to link $UX$. From eq. (35) we get

$$Q_U(x | \Theta_X^U, \Theta_{UX}^+) = 1 - \sum_{u=1}^{g_U} \pi_X(u | \Theta_{UX}^+) \sum_{x'=x+1}^{g_X} \theta_{x'}^u. \quad (41)$$

These expressions must be substituted into eq. (15). The assumptions of independence allow us to integrate over the parameters outside link $UX$, and after defining

$$\hat{Q}_V(x) \equiv 1 - \sum_{v=1}^{g_V} \hat{\pi}_X(v) \sum_{x'=x+1}^{g_X} \mu_{x'}^v, \quad (42)$$

and

$$R_U(x) \equiv \begin{cases} [\hat{\lambda}(x) - \hat{\lambda}(x+1)] \prod_V \hat{Q}_V(x) & \text{for } x < g_X \\ \hat{\lambda}(x) \prod_V \hat{Q}_V(x) & \text{for } x = g_X \end{cases} \quad (43)$$

we arrive at

$$P(\Theta_X^U | all \ cases)$$
$$\propto \left[ \sum_x R_U(x) \left( 1 - \sum_{u=1}^{g_U} \hat{\pi}_X(u) \sum_{x'=x+1}^{g_X} \mu_{x'}^u \right) \right] f(\Theta_X^U)$$
$$= \left[ \sum_x R_U(x) - \right.$$
$$\left. - \sum_{x'=1}^{g_X} \sum_{u=1}^{g_U} \theta_{x'}^u \left( \hat{\pi}_X(u) \sum_{x=0}^{x'-1} R_U(x) \right) \right] f(\Theta_X^U). \quad (44)$$

Finally, by comparing this expression to eq. (8) and substituting into eq. (13), we conclude that

$$\Delta_x^u = \cfrac{-(\sigma_x^u)^2 \, \hat{\pi}_X(u) \sum_{x'=0}^{x-1} R_U(x')}{\sum_{x'} R_U(x') - \sum_{x'=1}^{g_X} \sum_{u=1}^{g_U} \mu_{x'}^u \hat{\pi}_X(u) \sum_{x''=0}^{x'-1} R_U(x'')}. \quad (45)$$

In the case of binary variables, $g_U = g_X = 1$, and there is just one parameter $\theta_U^X$ for link $UX$. Using the notation $\lambda_i = \hat{\lambda}(X = i)$ and $\pi_X^U = \hat{\pi}_X(U = 1)$, the result becomes simplified to

$$\Delta_X^U = \frac{(\sigma_X^U)^2 \, \pi_X^U (\lambda_1 - \lambda_0)(1 - \pi_X^V \mu_X^V)}{\lambda_1 + (\lambda_1 - \lambda_0)(1 - \pi_X^U \mu_X^U)(1 - \pi_X^V \mu_X^V)}. \quad (46)$$

Besides repeating the same considerations as in the general case, we can also observe that, according to this last equation, when $\pi_X^U = 1$ (it means that $U$ is present), the evidential support for the presence of $X$ ($\lambda_1 > \lambda_0$) makes $\Delta_X^U$ positive, while $\Delta_X^U$ is negative for $\lambda_1 < \lambda_0$. This was the expected result, since parameter $\theta_X^U$ represents the probability that $U$ alone produces $X$.

# 4  CONCLUSIONS

This paper has presented a model for parameter adjustment in Bayesian networks. The starting point is a BN in which every conditional probability is given by its mean value and its standard deviation. The main virtue of this approach is that updating of parameters can be performed locally (distributed for every node), based on the $\pi$ and $\lambda$ messages of evidence propagation. The statistical model is cumbersome —more as a consequence of notation than of the ideas involved— but leads to simple algorithms. We tried to show the agreement between the results and what was expected from common sense.

We have given a mathematical definition of the generalized noisy OR–gate for multivalued variables and have shown how to compute probability in time proportional to the number of parents. In conjunction with local conditioning [2], this method can be used even in networks with loops, thus representing an important advantage over inference algorithms which work only on conditional probability tables. The learning model has also been applied to this gate.

The main shortcomings of this model reside in the strong assumptions of independence and in some approximation that might not be valid if standard deviations are wide and the observed evidence differs significantly from the expected values.

### Acknowledgements

The work was directed by Prof. José Mira as thesis advisor and supported by a grant from the Plan Nacional de Formación de Personal Investigador of the Spanish Ministry of Education and Science.

This paper has benefited from comments by Marek Druzdzel and the anonymous referees.